\pdfoutput=1

\documentclass[11pt]{article}

\usepackage{acl}

\usepackage{times}
\usepackage{latexsym}

\usepackage[T1]{fontenc}

\usepackage[utf8]{inputenc}

\usepackage{microtype}

\usepackage{booktabs}
\usepackage{multirow}
\usepackage{float}
\usepackage{graphicx}
\usepackage{amsmath}
\usepackage{url}
\usepackage{bm}

\def\smallcol{\hskip 6pt}
\def\tinycol{\hskip 2pt}

\title{Table Retrieval May Not Necessitate Table-specific Model Design}

\author{Zhiruo Wang, \quad Zhengbao Jiang, \quad Eric Nyberg, \quad Graham Neubig \\
Language Technologies Institute, Carnegie Mellon University \\ 
\texttt{\{zhiruow,zhengbaj,ehn,gneubig\}@cs.cmu.edu}}

\begin{document}
\maketitle
\begin{abstract}
Tables are an important form of structured data for both human and machine readers alike, providing answers to questions that cannot, or cannot easily, be found in texts. Recent work has designed special models and training paradigms for table-related tasks such as table-based question answering and table retrieval. Though effective, they add complexity in both modeling and data acquisition compared to generic text solutions and obscure which elements are truly beneficial. In this work, we focus on the task of table retrieval, and ask: ``is table-specific model design necessary for table retrieval, or can a simpler text-based model be effectively used to achieve a similar result?’’
First, we perform an analysis on a table-based portion of the Natural Questions dataset (NQ-table), and find that structure plays a negligible role in more than 70\% of the cases.
Based on this, we experiment with a general Dense Passage Retriever (DPR) based on text and a specialized Dense Table Retriever (DTR) that uses table-specific model designs. 
We find that DPR performs well without any table-specific design and training, and even achieves superior results compared to DTR when fine-tuned on properly linearized tables.
We then experiment with three modules to explicitly encode table structures, namely auxiliary row/column embeddings, hard attention masks, and soft relation-based attention biases. However, none of these yielded significant improvements, suggesting that table-specific model design may not be necessary for table retrieval.\footnote{The code and data are available at \url{https://github.com/zorazrw/nqt-retrieval}}
\end{abstract}

\section{Introduction}
\label{sec1:intro}
Tables are a valuable form of data that organize information in a structured way for easy storage, browsing, and retrieval~\citep{cafarella2008webtables,jauhar2016tables,zhang2020web}. They often contain data that is organized in a more accessible manner than in unstructured texts, or even not available in text at all~\citep{chen2020hybridqa}. Therefore, tables are widely used in question answering (QA)~\citep{pasupat2015compositional,zhong2017seq2sql,yu2018spider}. For open-domain QA, the ability to retrieve relevant tables with target answers is crucial to the performance of end-to-end QA systems~\citep{herzig2021open}. For example, in the Natural Questions~\citep{kwiatkowski2019natural} dataset, 13.2\% of the answerable questions can be addressed by tables and 74.4\% by texts. 

\begin{figure}
    \centering
    \includegraphics[width=7.7cm]{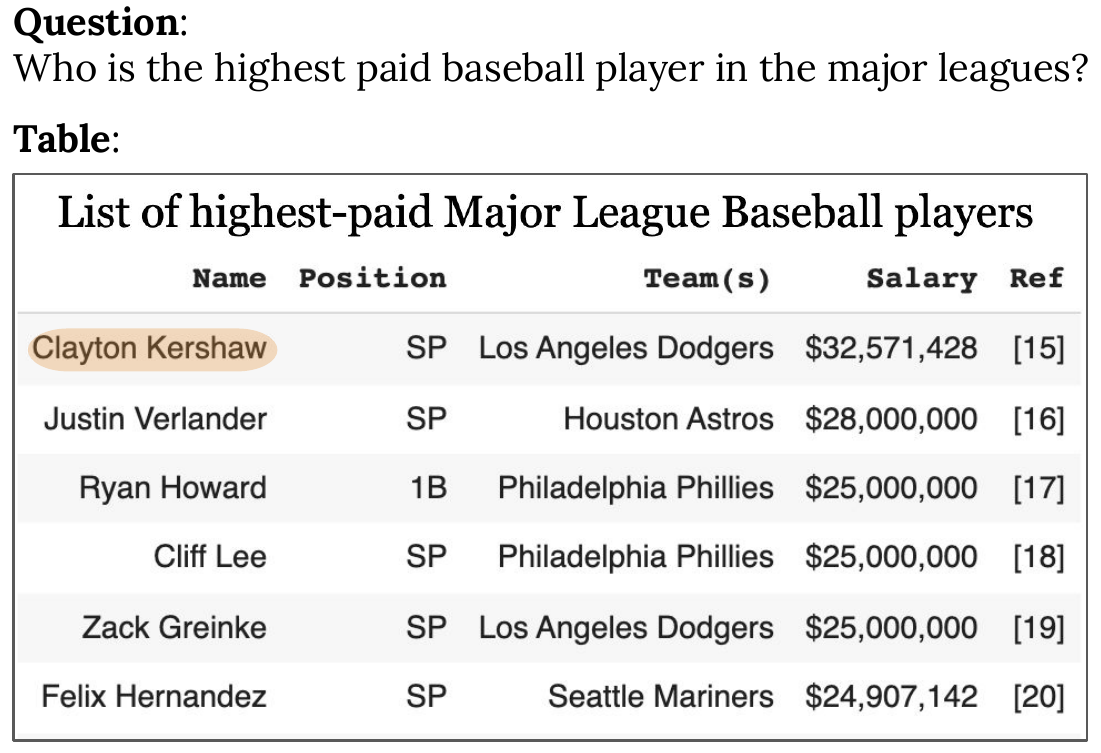}
    \caption{A correct table can be identified by matching key phrases in question to those in the table title and header cells. }
    \label{fig:example-intro}
\end{figure}

Because tables are intuitively different from unstructured text, most previous works consider text-based methods to be functionally incapable of processing tables effectively and create special-purpose models with table-specific architectures and training methods, adding auxiliary structure-indicative parameters~\citep{herzig2020tapas,wang2021tuta,deng2020turl,yang2022tableformer}, enforcing structure-aware attention~\citep{yin2020tabert,wang2021tuta,zayats2021representations}, and table-oriented pre-training objectives~\citep{deng2020turl,yin2020tabert,wang2021tuta,liu2021tapex,yu2020grappa}. 
Though effective in many tasks, these special-purpose models are more complex than generic solutions for textual encoding, and must be intentionally built for and trained on tabular data. In addition, because these methods modify both the model design and the training data, it is difficult to measure the respective contributions of each of these elements.

Particularly for question-based table retrieval, we hypothesize that content matching is paramount, and little, if any, structural understanding may be required.
For example, given a question ``Who is the highest paid baseball player in the major leagues?'' in~\autoref{fig:example-intro}, a correct table can be retrieved by simply identifying the phrase ``highest-paid'', ``major league'', and ``baseball player'' in the table title, and matching the semantic type of ``Who'' to the ``Name'' header.
Hence, any benefit demonstrated by table-based models may well come from good training data while table-specific model design has a limited influence. 

In this paper, we specifically ask: ``Does table retrieval require table-specific model design, or can properly trained generic text retrievers be exploited to achieve similar performance with less added complexity?'' Our work centers around the table-based open-domain QA dataset, NQ-table~\citep{herzig2021open}, a subset of the Natural Questions (NQ) dataset~\citep{kwiatkowski2019natural} where each question can be answered by part(s) of a Wikipedia table. 
We start with manual analysis of 100 random samples from NQ-table and observe that consideration of table structure seems largely unnecessary in over 70\% of the cases, while the remaining 30\% of cases only require simple structure understanding such as row/column alignment without structure-dependent complex reasoning chains (\autoref{sec:dataset}). 
With this observation, we experiment with two strong retrieval models: a general-purpose text-based retriever (DPR; \citet{karpukhin2020dense}) and a special-purpose table-based retriever (DTR; \citet{herzig2021open}).
We find that DPR, without any table-specific model design or training, achieves similar accuracy as the state-of-the-art table retriever DTR, and further fine-tuning on NQ-table yields significantly superior performance, casting doubt on the necessity of table-specific model design in table retrieval (\autoref{sec:retrievers}).
Using DPR as the base model, we then thoroughly examine the effectiveness of both encoding structure implicitly with structure-preserving table linearization (\autoref{sec:ablation}) and encoding structure explicitly with table-specific model design, such as auxiliary embeddings and specialized attention mechanisms (\autoref{sec:finetune}).
We find that models can already achieve a degree of structure awareness using properly linearized tables as inputs, and additionally adding explicit structure encoding model designs does not yield a further improvement.
In sum, the results reveal that a strong text-based model is competitive for table retrieval, and table-specific model designs may have limited additional benefit.
This indicates the potential to directly apply future improved text retrieval systems for table retrieval, a task where they were previously considered less applicable.

\section{NQ-table Analysis: How Much Structure Does Table Retrieval Require?}
\label{sec:dataset}

The NQ-table dataset~\citep{herzig2021open} is a subset of the Natural Questions (NQ) dataset~\citep{kwiatkowski2019natural} which contains questions from real users that can be answered by Wikipedia articles. Previous works on text-based QA extract the text portion from source Wikipedia articles that can answer around 71k questions, while NQ-table extract tables that contain answers for 12k questions.
Unless otherwise specified, we use NQ-text to denote the commonly referred NQ dataset that can be answered by texts.


To better understand to what extent (if any) is structure understanding required by table retrieval, we perform a manual analysis on the NQ-table dataset. Specifically, we randomly sample 100 questions and their relevant tables then categorize their matching patterns.

\begin{figure}
    \centering
    \includegraphics[width=7.8cm]{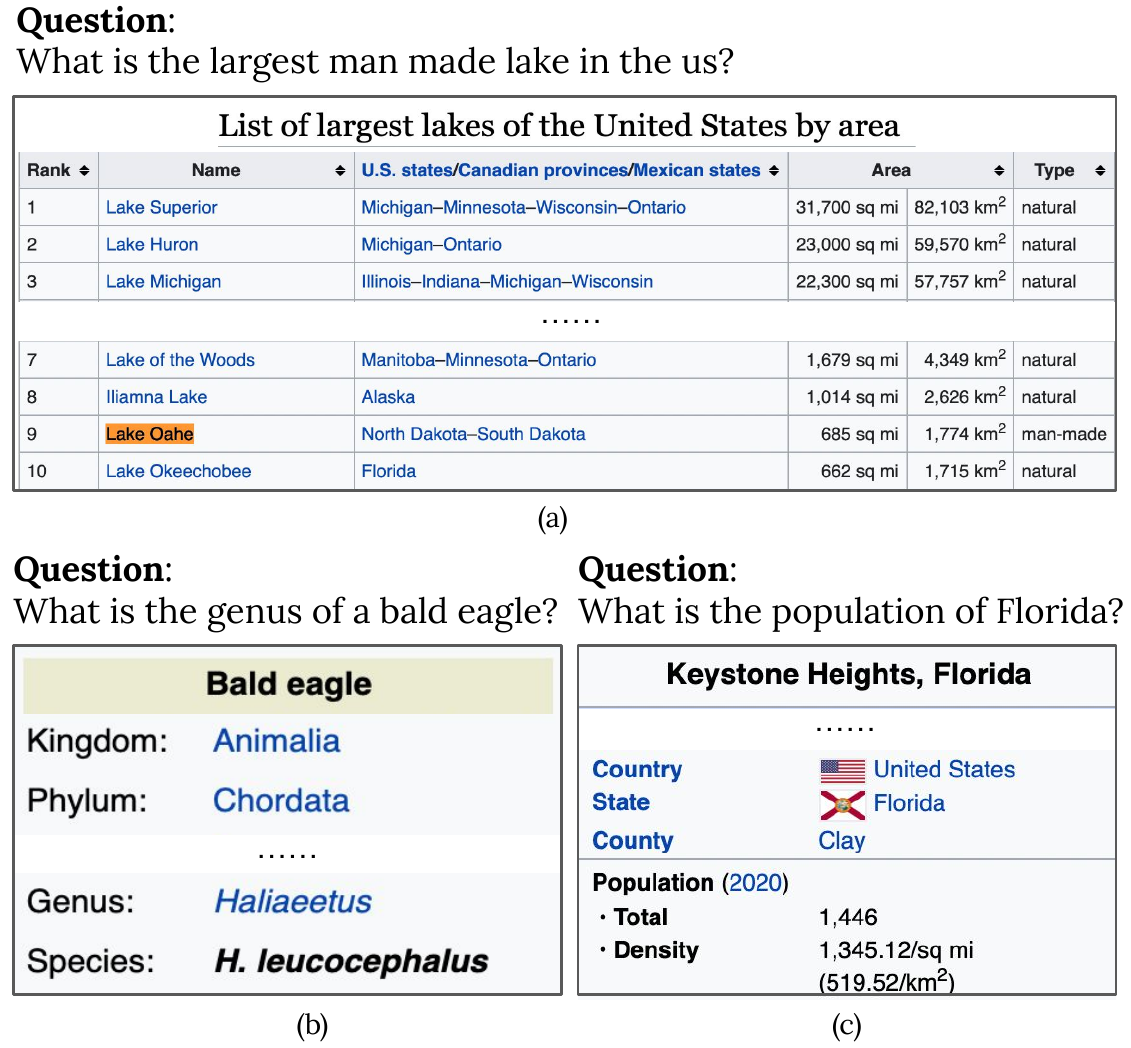}
    \caption{Table (a) matches the question by its title, (b) matches topic in title and answer type in header, and in (c) knowing the column alignment helps.}
    \label{fig:nqt}
\end{figure}

\paragraph{Keyword Matching Without Structural Concern} 
Aligning with the insight that retrieval often emphasizes content matching rather than complex reasoning~\citep{rogers2021qa}, we find that 71 out of the 100 samples only require simple keyword matching, where 18 questions fully match with table titles (\autoref{fig:nqt} (a)) and the other 53 questions further match with table headers (\autoref{fig:nqt} (b)).

\paragraph{Retrieval that Requires Row/Column Alignment}

For the other 29 samples, understanding table structure is helpful but only simple row/column alignment is needed. 
21 of them require locating content cells in a specific column and combining the information from headers. For example in~\autoref{fig:nqt}(c), under the general header ``Population'', one should locate the ``Total'' field by their structural relation to confirm that the `total number' measure of `population' exists. 
In addition, 7 of the samples are somewhat ambiguous and may require external knowledge or question clarification~\citep{min2020ambigqa}.

In summary, our analysis reveals that understanding table structure is not necessary in the majority of cases, and even for cases where structural information is useful, they merely require aligning the rows/columns instead of building complex chains of reasoning.

\section{Text Retrieval vs Table Retrieval}
\label{sec:retrievers}

Given the previous analysis, we hypothesize that general-purpose text-based retrievers without table-specific designs might not be necessarily worse than special-purpose table-based retrievers, contradictory to what most previous work has assumed \cite{herzig2021open,herzig2020tapas,yin2020tabert,wang2021tuta}.
Properly trained text-based retrievers might even outperform table-based retrievers because the strong content matching ability learned on text retrieval datasets can transfer to the table retrieval task.

To validate these assumptions, we examine two representative retrieval systems: the text-based Dense Passage Retriever (DPR) and the table-based Dense Table Retriever (DTR). 
We first briefly introduce their input formats and model architectures (\autoref{sec:dpr},\autoref{sec:dtr}), then conduct experiments in both zero-shot and fine-tuning settings and compare their table retrieval performance (\autoref{sec:simple_ft}).

\subsection{Text Retriever: DPR}
\label{sec:dpr}

We choose DPR~\cite{karpukhin2020dense} as a representative text retrieval model, mainly because of (1) its impressive performance across many text-related retrieval tasks, and (2) its similarity with DTR from both training and modeling perspectives, which make it easy to make fair comparisons.

DPR comprises a question-context bi-encoder built on BERT~\cite{devlin2018bert}, which includes three types of input embeddings as summarized in~\autoref{tab:compare-embeddings}.
The question encoder $\text{BERT}_q$ encodes each question $\bm{q}$ and outputs its dense representation using the representation of $\text{[CLS]}$ token, denoted as $\bm{h}_{\bm{q}} = \text{BERT}_q(\bm{q})\text{[CLS]}$. 
The context encoder works similarly. To enable tables for sequential context inputs, we linearize each table into a token sequence $\bm{T}$, which is then fed into the context encoder $\text{BERT}_c$ to obtain its dense representation $\bm{h}_{T} = \text{BERT}_c(T)\text{[CLS]}$. 
The similarity score between a question $\bm{q}$ and a table $\bm{T}$ is computed as the dot product of two vectors $sim(\bm{q}, T) = \bm{h}_{\bm{q}} \cdot \bm{h}_{T}$. 

DPR has been trained only on sequential text contexts. For each question in the NQ-text training set, the model is trained to select the correct context that contains the answer from a curated batch of contexts including both the annotated correct contexts and mined hard negative contexts. 

To convert tables into the DPR input format, we linearize tables into token sequences. We concatenate the title, the header row, and subsequent content rows using a period `.' (row delimiter). Within each header or content row, we concatenate adjacent cell strings using a vertical bar `|' (cell delimiter). A template table linearization reads as $[\text{title}] . [\text{header}]. [\text{content}_1] . ~\cdots~ . [\text{content}_n]$.
Although the BERT encoder has the capacity for a maximum of 512 tokens, DPR is only exposed to contexts no longer than 100 words during training and testing. To avoid potential discrepancies between its original training and our inference procedure, we shorten long tables by selecting the first few rows that fit into the 100-word window. 

\subsection{Table Retriever: DTR}\label{sec:dtr}

Dense Table Retriever (DTR) \cite{herzig2021open} is the current state-of-the-art table retrieval model on the NQ-table dataset. 

\paragraph{Model Architecture}
DTR largely follows the bi-encoder structure of DPR, but differs from it in the embedding layer. 
As shown in~\autoref{tab:compare-embeddings}, DTR utilizes the existing embeddings in alternative ways and introduces new types of embeddings specifically designed to encode tables. 

Both models use the BERT vocabulary index for token embedding. For the segment index, DPR assigns all tokens in a sequence to index $0$, while DTR distinguishes the title from table content by assigning $0$ and $1$, respectively. For positions, DPR inherits from BERT the sequence-wise order index $[0, 1, 2, ..., \text{sequence length} - 1]$; DTR adopts a cell-wise reset strategy that records the index of a token within its located cell $[0, 1, ..., \text{cell length}-1]$. 

Most importantly, DTR introduces row and column embeddings to encode the structural position of each token in the cell that it appears. This explicit join of three positional embeddings is potentially more powerful than the BERT-style flat index. Besides, concerning the high frequency of numerical values in tables, DTR adds a ranking index for each token if it is part of a number. 

\begin{table}[H]

\centering
\resizebox{0.49\textwidth}{!}{
    \begin{tabular}{l|c|c}
    \toprule
    \multicolumn{1}{c|}{\textbf{Embeddings}} & \textbf{DPR} & \textbf{DTR} \\
    \midrule
    \textbf{token} & {BERT vocab} & {BERT vocab} \\
    \textbf{segment} & {0 for all tokens} & {0 for text, 1 for table} \\
    \textbf{position} & {sequential} & {cell-wise reset} \\
    \textbf{row} & {-} & {row index} \\
    \textbf{column} & {-} & {column index} \\
    \textbf{rank} & {-} & {rank of token value} \\
    \bottomrule
    \end{tabular}
}
\caption{Comparison of DPR and DTR embeddings. }
\label{tab:compare-embeddings}

\end{table}

\paragraph{Training Process}

\begin{figure}[h]
    \centering
    \includegraphics[width=7.7cm]{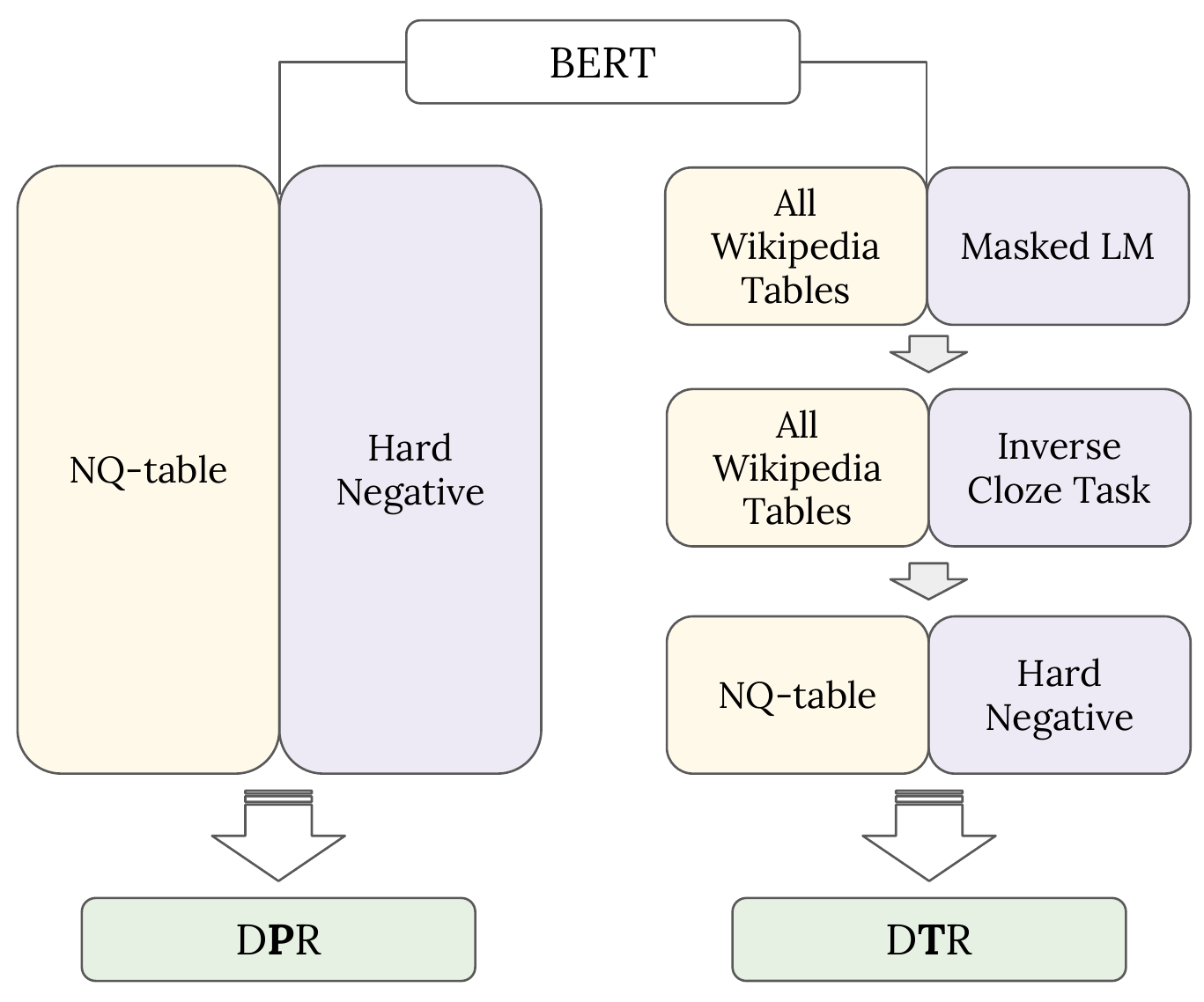}
    \caption{Comparison of DPR and DTR training. }
    \label{fig:training-process}
\end{figure}

DTR also has a more complex training process than DPR. As summarized in~\autoref{fig:training-process}, DTR has a three-stage training using tables. 

First, model parameters, except for those extra table-specific embeddings, are initialized with BERT weights. The model is then pre-trained on all Wikipedia tables using the Masked LM (MLM)~\citep{devlin2018bert} task, yielding the TAPAS~\cite{herzig2020tapas} model. Second, to leverage TAPAS to the retrieval task, it is further pre-trained using the Inverse Cloze task (ICT) introduced by ORQA~\cite{lee2019latent}, again, on all Wikipedia tables. 
Third, the model is trained on the specific NQ-table dataset, similar to the way that DPR is trained on text retrieval datasets: for each question in the NQ-table training set, DTR uses the annotated table as the positive context and self-mined tables without answers as hard negative (HN) contexts.

\subsection{Text Retrieval Benefits Table Retrieval}\label{sec:simple_ft}

To evaluate the benefit on table retrieval from training on in-domain text retrieval datasets, we compare the performance of DPR and BERT \citep{devlin2018bert} after fine-tuning on NQ-table.

As shown in \autoref{tab:retrieval}, \emph{BERT-table} significantly underperforms \emph{DPR-table}, indicating that training on in-domain text retrieval datasets benefits the table retrieval task.
We conjecture that the large gap is essentially because (1) NQ-text and NQ-table questions share similar characteristics hence are agnostic to the format of answer source~\citep{wolfson2020break}, and (2) NQ-text has a larger size than NQ-table (71k versus 12k).

\subsection{DPR vs DTR}\label{sec:notnecessary}

To verify if table-specific model designs in DTR are necessary, we start with comparing the original DPR with DTR to evaluate their off-the-shelf performance, then proceed to fine-tune DPR on NQ-table to examine the how much improvement can be brought by training data. We evaluate both models on NQ-table test set and measure the retrieval accuracy by computing the portion of questions where the top-k retrieved tables contain the answer.

For DPR experiments, we use the latest published checkpoint\footnote{\url{https://github.com/facebookresearch/DPR}} where the hard-negative text passages are mined using the DPR checkpoint saved in the previous round. 
To reproduce the DTR performance, we use the published checkpoints\footnote{\url{https://github.com/google-research/tapas/blob/master/DENSE_TABLE_RETRIEVER.md}} and run the retrieval inference. 

To curate training samples for questions in the NQ-table training set, we take the same positive table used in DTR training. For negative contexts, we use the original DPR checkpoint to retrieve the top-100 table candidates for each question, from which we take the highest-ranked tables without answers as the hard negatives. 
We train with a batch size of $16$ and a learning rate of $2\mathrm{e}{-5}$. Experiments are finished on four NVIDIA Tesla V100 GPUs.

Note that the published DPR and DTR checkpoints are not strictly comparable, since the size of DPR base falls between the DTR medium and DTR large with respect to the number of parameters. We report the performance of DTR in both medium and large size to approximate the lower and upper bounds for the DTR base model. 
\begin{table}[h]
    \centering
    \resizebox{0.5\textwidth}{!}{
    \begin{tabular}{c|ccc}
        \toprule
        \textbf{Size} & \textbf{Layers} & \textbf{Attention Heads} & \textbf{Hidden Size}\\ 
        \midrule
        {medium} & {8} & {8} & {512} \\
        {base} & {12} & {8} & {768} \\
        {large} & {24} & {16} & {1024} \\
        \bottomrule
    \end{tabular}
    }
\caption{Hyper-parameters for BERT models of varied sizes. Models of different sizes vary in the number of transformer layers, the number of heads in the self-attention module, and the dimension of hidden states.}
\label{tab:model-size}
\end{table}

\autoref{tab:model-size} shows the configurations of BERT-variants in different sizes. 
As can be seen from the hyper-parameter values, models of medium size have the smallest capacity, base is an intermediate configuration, and large size is the biggest.

As reported in \autoref{tab:retrieval}, DPR is able to achieve a zero-shot retrieval accuracy (\emph{DPR}) on NQ-table that is fairly close to the state-of-the-art DTR model, even without any table-specific model design and training. 
Further, simply fine-tuning DPR on NQ-table (\emph{DPR-table}) using the same annotated positive and mined hard-negative tables as DTR increases the performance by a large margin, achieving superior performance than DTR, especially at top ranking positions (i.e., small k). 

\begin{table}[H]
\centering
\small
    \begin{tabular}{@{}c|c@{\smallcol}c@{\smallcol}c@{\smallcol}c@{\smallcol}c@{\smallcol}c@{}}
    \toprule
    \multirow{2}{*}{\textbf{Model}} & \multicolumn{5}{c}{\textbf{Retrieval Accuracy}} \\
    {} & \textbf{@1} & \textbf{@5} & \textbf{@10} & \textbf{@20} & \textbf{@50} \\
    \midrule
    \multicolumn{1}{l|}{DTR (medium)} & {62.32} & {82.51} & {86.75} & {91.51} & {94.26} \\
    \multicolumn{1}{l|}{DTR (large)} & {63.98} & {84.27} & \textbf{89.65} & \textbf{93.48} & \textbf{95.65} \\
    \midrule
    \multicolumn{1}{l|}{BERT-table} & {60.97} & {79.81} & {85.51} & {88.20} & {91.62} \\
    \midrule
    \multicolumn{1}{l|}{DPR} & {57.04} & {80.54} & {86.13} & {89.54} & {92.34} \\
    \multicolumn{1}{l|}{DPR-table} & \textbf{67.91} & \textbf{84.89} & {88.72} & {90.58} & {92.86} \\
    \bottomrule
    \end{tabular}
\caption{Top-k table retrieval accuracy on NQ-table test set. \emph{DPR} is the original model checkpoint. \emph{DPR-table} and \emph{BERT-table} are DPR and BERT fine-tuned on NQ-table respectively.}
\label{tab:retrieval}
\end{table}

These observations question the necessity of both table-specific model designs listed in \autoref{tab:compare-embeddings} and table-specific pre-training listed in \autoref{fig:training-process}.
Given the task analysis in \autoref{sec:dataset} that table retrieval only requires simple structure understanding, we hypothesize that DPR, trained with table inputs linearized from top-to-bottom and left-to-right, is functionally capable of implicitly encoding simple table structure such as row/column alignment, and the benefit of extra table-specific model designs is minimal.
To thoroughly and rigorously verify our hypothesis, we first examine the effect of different ordering in table linearization in \autoref{sec:ablation}, then experiment with three widely-used structure injection model designs by adding them on DPR in \autoref{sec:finetune}.

\section{Implicit Structure Encoding from Linearized Tables}
\label{sec:ablation}

The simplest way to encode table structure is to linearize the table following the top-to-bottom left-to-right order and insert delimiters between cells and rows, from which the sequence-oriented transformer models should also be able to recover the two-dimensional table structure.

We hypothesize that this type of implicit structure encoding is sufficient for table retrieval, which only requires simple structure understanding.
To verify this, we manipulate linearized tables by randomly shuffling their rows/columns (\autoref{sec:shuffle}) or removing the delimiters (\autoref{sec:del}), and examine how these perturbation affect the final performance. 

\subsection{Shuffling Rows and Columns}\label{sec:shuffle}
Our first experiment focuses on the order of table linearization: if DPR relies on a proper linearization to capture table structure, randomly shuffling the table contents should corrupt the structure information and hurt the representation quality, leading to lower retrieval accuracy. 

To verify this, we shuffle table cells within each \emph{row}, each \emph{column}, or \emph{both}. 
Cells in the same row often describe the same entity from multiple properties according to their column headers, therefore shuffling the order of multiple cells in the same row corrupts their alignment with header cells.
Meanwhile, cells in the same column are often of the same semantic type but are attributes to different entities in different rows, shuffling the order of cells in the same column breaks their alignment with entities. 
We also examine shuffling on both dimensions, which completely removes the order information from the table linearizations.

Since models trained on properly linearized tables might be prone to the train-test discrepancy when tested on shuffled tables, we conjecture that the gap between testing on proper tables and shuffled tables cannot be fully attributed to the loss of order information.
We therefore conduct a more rigorous experiment by fine-tuning DPR on shuffled tables in both dimensions (\emph{DPR-table w/ shuffle}) and test it on both proper and shuffled tables. 

\begin{table}[H]
\small
\centering
\begin{tabular}{@{}l@{\tinycol}|l@{\tinycol}|c@{\smallcol}c@{\smallcol}c@{\smallcol}c@{\smallcol}c@{}}
\toprule
\multicolumn{2}{c|}{\textbf{Method}} & \multicolumn{5}{c}{\textbf{Retrieval Accuracy}} \\
\multicolumn{1}{c}{\textbf{Model}} & \multicolumn{1}{c|}{\textbf{Shuffle}} & \textbf{@1} & \textbf{@5} & \textbf{@10} & \textbf{@20} & \textbf{@50} \\
\midrule
\multirow{4}{*}{DPR} & \multicolumn{1}{c|}{-} & {57.04} & {80.54} & {86.13} & {89.54} & {92.34} \\
\cmidrule{2-7}
{} & {row} & {55.18} & {79.19} & {85.82} & {89.75} & {92.44} \\
{} & {column} & {57.04} & {80.85} & {86.65} & {89.34} & {92.55} \\
{} & {both} & {57.97} & {79.61} & {84.89} & {89.44} & {92.55} \\
\midrule 
\multirow{4}{*}{DPR-table} & \multicolumn{1}{c|}{-} & {67.91} & {84.89} & {88.72} & {90.58} & {92.86} \\
\cmidrule{2-7}
{} & {row} & {55.18} & {76.09} & {80.64} & {85.40} & {89.23} \\
{} & {column} & {58.39} & {77.74} & {82.92} & {86.44} & {89.86} \\
{} & {both} & {54.76} & {75.16} & {80.64} & {84.87} & {88.82} \\
\midrule 
 & \multicolumn{1}{c|}{-} & {62.94} & {80.12} & {84.99} & {88.92} & {91.30} \\
\cmidrule{2-7}
{DPR-table} & {row} & {62.11} & {80.95} & {85.30} & {88.82} & {91.72} \\
{w/ shuffle} & {column} & {64.91} & {82.30} & {86.75} & {89.54} & {92.55} \\
{} & {both} & {63.35} & {81.06} & {85.20} & {89.34} & {91.93} \\
\bottomrule
\end{tabular}
\caption{Top-k table retrieval accuracy on shuffled NQ tables, using the original \emph{DPR}, the fine-tuned (\emph{DPR-table}), and the fine-tuned on shuffled tables (\emph{DPR-table w/ shuffle}).}
\label{tab:perturb}
\end{table}

As shown in~\autoref{tab:perturb}, on the original DPR model, all table shuffling strategies result in minor variations in retrieval accuracy, which is intuitive because DPR has never been trained on linearized tables and it is not sensitive to cell orders. 

The performance of fine-tuned DPR (\emph{DPR-table}) drops significantly when tested on shuffled tables, similar to the previous finding that T5 model is also sensitive to the ordering of structured knowledge~\citep{xie2022unifiedskg}. Besides the potential discrepancy in table layout between training and test inputs, this may indicate that DPR, although without explicit structure encoding modules, also learns to implicitly capture structures by training on linearized table inputs.

To ablate out the influence of train-test discrepancy, we also fine-tune DPR on shuffled positive and negative tables. As expected, \emph{DPR-table w/ shuffle} does not suffer from train-test discrepancy.
While DPR fine-tuned on shuffled tables still outperforms the original DPR (57.04$\rightarrow$62.94@1), the improvement is not as significant as the improvement obtained by fine-tuning on proper tables (57.04$\rightarrow$67.91@1), indicating that DPR is able to utilize structure-preserving table linearizations to encode structures during training. 

Comparing different shuffling dimensions, we notice that in-\emph{row} shuffling hurts the performance more than in-\emph{column} shuffling, indicating that preserving semantic type alignment within each column is more important than preserving entity alignment within each row for table retrieval.

\subsection{Removing Delimiters Between Rows/Cells}\label{sec:del}

In this section, we study the impact of delimiters in helping models to encode table structures. If delimiters are not included, it is theoretically impossible to recover the table structure even from properly linearized tables, because the boundaries between different cells and rows are unknown. To verify if delimiters can serve as effective indicators of table structure, we study the usefulness of both inserting delimiter (`|') between cells and inserting delimiter (`.') between rows. 

Similarly to the previous experiment, we evaluate (1) the original DPR model (\emph{DPR}), (2) the DPR fine-tuned on tables with delimiters (\emph{DPR-table}), and (3) the one fine-tuned on linearized tables without delimiters (\emph{DPR-table w/o delimiter}).

\begin{table}[H]
\small
\centering
\resizebox{0.5\textwidth}{!}{
    \begin{tabular}{@{}l@{\tinycol}|l@{\tinycol}|c@{\smallcol}c@{\smallcol}c@{\smallcol}c@{\smallcol}c@{}}
    \toprule
    \multicolumn{2}{c|}{\textbf{Method}} & \multicolumn{5}{c}{\textbf{Retrieval Accuracy}} \\
    \multicolumn{1}{c}{\textbf{Model}} & \multicolumn{1}{c|}{\textbf{Delimiter}} & \textbf{@1} & \textbf{@5} & \textbf{@10} & \textbf{@20} & \textbf{@50} \\
    \midrule
    \multirow{4}{*}{DPR} & all & {57.04} & {80.54} & {86.13} & {89.54} & {92.34} \\
    \cmidrule{2-7}
    {} & {cell} & {56.00} & {79.40} & {85.30} & {89.54} & {92.34} \\
    {} & {row} & {54.24} & {77.54} & {82.92} & {87.68} & {92.13} \\
    {} & {none} & {55.49} & {79.09} & {84.78} & {89.44} & {92.03} \\
    \midrule 
    \multirow{4}{*}{DPR-table} & all & {67.91} & {84.89} & {88.72} & {90.58} & {92.86} \\
    \cmidrule{2-7}
    {} & {cell} & {55.80} & {75.26} & {81.16} & {85.20} & {89.23} \\
    {} & {row} & {55.07} & {74.95} & {80.75} & {84.68} & {89.65} \\
    {} & {none} & {56.63} & {76.19} & {81.26} & {86.13} & {89.75} \\
    \midrule 
     & all & {63.46} & {81.47} & {85.09} & {88.82} & {92.13} \\
    \cmidrule{2-7}
    {DPR-table} & {cell} & {63.04} & {83.02} & {87.47} & {90.06} & {92.13} \\
    {w/o delimiter} & {row} & {63.35} & {80.54} & {85.20} & {89.34} & {92.03} \\
    {} & {none} & {64.49} & {81.88} & {86.23} & {89.86} & {92.55} \\
    \bottomrule
    \end{tabular}
}
\caption{NQ-table retrieval accuracy with linearized table w/ and w/o cell and row delimiters. \emph{cell} linearizes table by only inserting delimiters between cells, \emph{row} only inserts delimiters between rows, and \emph{none} inserts neither.}
\label{tab:delimiter}
\end{table}

As shown in~\autoref{tab:delimiter}, for \emph{DPR}, although the overall performance drop is small without delimiters, separating cells is more important than separating rows, which is intuitive because the number of cells is larger than the number of rows.
On \emph{DPR-table} that learns from properly delimited tables, the influence is more significant, and the extent of dropping is similar to that of table structure shuffling in \autoref{tab:perturb}.
Also similar to the previous findings, training on non-delimited tables (\emph{DPR-table w/o delimiter}) improves over the original DPR, but the improvement is not as significant as the improvement obtained by fine-tuning on delimited tables, suggesting that cell and row delimiters help models encode table structure.

\section{Explicit Structure Encoding with Table-specific Model Design}
\label{sec:finetune}

From the previous section, we conclude that DPR can already encode simple table structures based on structure-preserving linearized tables with correct cell order and delimiters. The next question is ``can explicit table-specific model designs encode more complex structure that is useful beyond the capacity of implicit encoding?''

\begin{figure*}
    \includegraphics[width=16.0cm]{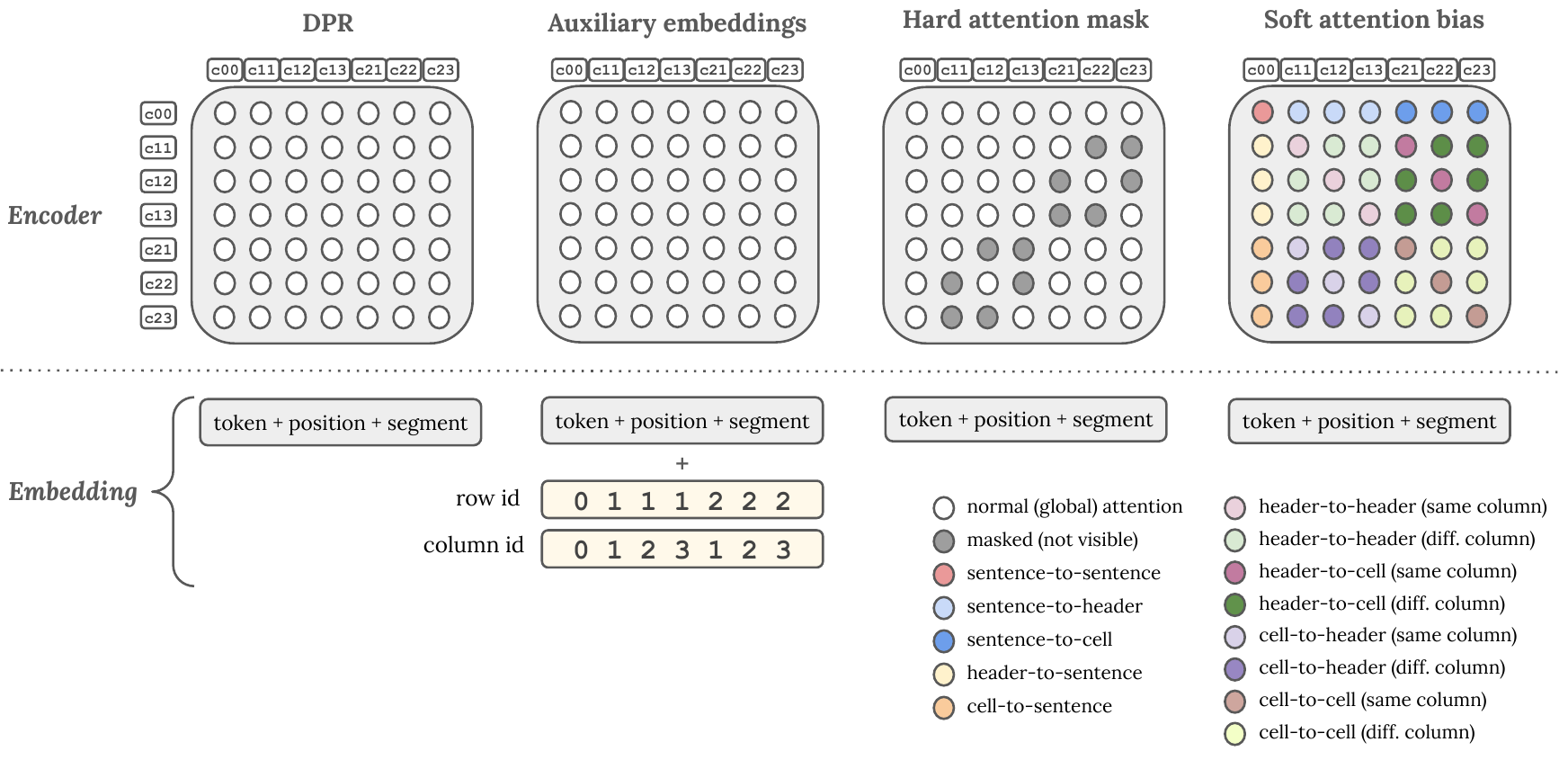}
    \caption{Illustration of three explicit structure encoding methods.}
    \label{fig:structure-injection}
\end{figure*}

In this section, we examine three widely used table-specific modules to explicitly encode table structure information by adding these modules on top of the DPR architecture.
As summarized in \autoref{tab:structure-methods} and illustrated in \autoref{fig:structure-injection}, we categorize existing methods for table-specific structure encoding into three representative types: (1) auxiliary table-specific embeddings, (2) restricted hard attention mask to enforce structure-aware attention, and (3) soft attention bias based on the structural relations of cell pairs. 
For each component, we add it onto the DPR architecture and fine-tune under the same setting as for \emph{DPR-table}.


\begin{table}[H]
\small
\centering
    \begin{tabular}{@{}l|l@{\smallcol}c@{}}
    \toprule
    \multicolumn{1}{c|}{\textbf{Method}} & \multicolumn{1}{c}{\textbf{Papers}} \\
    \midrule
    \multirow{4}{*}{auxiliary embeddings} & {TAPAS~\citep{herzig2020tapas}} \\
    {} & {MATE~\citep{eisenschlos2021mate}} \\
    {} & {TUTA~\citep{wang2021tuta}} \\
    {} & {TABBIE~\citep{iida2021tabbie}} \\
    \midrule 
    \multirow{6}{*}{hard attention mask} & {TURL~\citep{deng2020turl}} \\
    {} & {SAT~\citep{zhang2020table}} \\
    {} & {ETC~\citep{ainslie2020etc}} \\
    {} & {DoT~\citep{krichene2021dot}} \\
    {} & {MATE~\citep{eisenschlos2021mate}} \\
    {} & {TUTA~\citep{wang2021tuta}} \\
    \midrule 
    \multirow{2}{*}{soft attention bias} & {RAT-SQL~\citep{wang2020rat}} \\
    `{} & {TableFormer~\citep{yang2022tableformer}} \\
    \bottomrule
    \end{tabular}
\caption{Structure encoding methods used in previous works for table-related tasks.}
\label{tab:structure-methods}
\end{table}

\subsection{Auxiliary Row and Columns Embeddings}
We first examine if adding table-specific embedding parameters would bring additional improvement. Specifically, we add row and column embeddings into the DPR to encode the row and column indices of tokens, which is denoted as \emph{DPR-table w/ emb}. Both row and column indices are 1-indexed, and 0 is used for tokens that are not part of the table (e.g., title). 
We initialize row/column embeddings with zero to allow smooth continual learning.

\subsection{Hard Attention Mask}
Another approach is to enforce structure-aware attention using hard attention mask that only allows attention between elements within their mutual structural proximity, with the assumption that elements are only semantically relevant to elements in their structural proximity. 
Specifically, \citet{krichene2021dot,eisenschlos2021mate,deng2020turl} sparsify the attention mask such that each token is only visible to other tokens that are either within the same row or the same column. 
We apply this masking strategy when fine-tuning DPR and denote this setting as \emph{DPR-table w/ mask}.

\subsection{Soft Relation-based Attention Bias}
The third method is to bias the attention weight between two tokens based on their structural relation, which is a more fine-grained way to enforce structure-aware attention than hard mask. Specifically, different bias scalars are added to the attention scores based on the relation between two cells. \citet{wang2020rat} categorize relations by columns, while \citet{yang2022tableformer} defines 13 relations based on which component the token belongs to: sentence, header, and cell. A more concrete example is illustrated in \autoref{fig:structure-injection}. Relational bias is invariant to the numerical indices of rows and columns, which is more robust to answer-invariant structure perturbation.
We follow \citet{yang2022tableformer} to add soft attention bias on DPR with 13 relations.

\subsection{Results and Analysis}

As shown in~\autoref{tab:structure}, methods that explicitly encode table structures, either with additional row/column embeddings (\emph{w/ emb}), hard attention mask (\emph{w/ mask}), or soft relation-based attention bias (\emph{w/ bias}), do not bring improvements over the \emph{DPR-table} baseline, indicating that given the capacity of DPR in implicitly encoding structure from linearized tables, the benefit of using special-purpose structure encoding modules is minimal.

\begin{table}[H]
\centering
\small
    \begin{tabular}{@{}c|c@{\smallcol}c@{\smallcol}c@{\smallcol}c@{\smallcol}c@{\smallcol}c@{}}
    \toprule
    \multirow{2}{*}{\textbf{Model}} & \multicolumn{5}{c}{\textbf{Retrieval Accuracy}} \\
    {} & \textbf{@1} & \textbf{@5} & \textbf{@10} & \textbf{@20} & \textbf{@50} \\
    \midrule
    \multicolumn{1}{l|}{DPR-table} & \textbf{67.91} & \textbf{84.89} & \textbf{88.72} & \textbf{90.58} & {92.86} \\
    \midrule
    \multicolumn{1}{l|}{\quad w/ emb} & {65.73} & {81.99} & {86.02} & {89.23} & {92.86} \\
    \multicolumn{1}{l|}{\quad w/ mask} & {62.11} & {81.88} & {86.96} & {89.86} & \textbf{93.06} \\
    \multicolumn{1}{l|}{\quad w/ bias} & {65.42} & {82.23} & {86.75} & {89.54} & {92.13} \\
    \bottomrule
    \end{tabular}
\caption{Top-k table retrieval accuracy on NQ-table test. \emph{DPR-table} is fine-tuned on the NQ-table without any table-specific modules, while the other three methods add auxiliary row/column embeddings (\emph{w/ emb}), hard attention mask (\emph{w/ mask}), and soft relation-based attention bias respectively (\emph{w/ bias}).}
\label{tab:structure}
\end{table}


\section{Related Work} 
\paragraph{Open-domain Question Answering} 
Open-domain QA systems often use a retriever-reader pipeline, where the retriever retrieves relevant contexts and the reader extracts or generates answer from them. Because the candidate context corpus is usually large with millions of documents, good retrieval accuracy is crucial for open-domain QA systems~\citep{karpukhin2020dense}. 
Beyond texts, another common sources for answering open-domain questions is tables. ~\citet{herzig2021open} recently identified a subset of Natural Questions (NQ) dataset~\citep{kwiatkowski2019natural} that is answerable by Wikipedia tables. ~\citet{oguz2021unik} found that incorporating structured knowledge is beneficial for open-domain QA tasks. ~\citet{ma2021open} showed that verbalizing structured knowledge into fluent text bring further gains over raw format for open-domain QA. Different from prior work, our paper analyzes different strategies for encoding tables with a focus on the task of table retrieval. 

\paragraph{Table Understanding} 
To encode the relational structure of tables, CNNs~\citep{chen2019colnet}, RNNs~\citep{gol2019tabular}, LSTMs~\citep{fetahu2019tablenet}, and their combinations~\citep{chen2019learning} are explored. In addition, graph neural networks (GNN) are used, especially for tables with complex structures~\cite{koci2018table,zayats2021representations,vu2021graph,bhagavatula2015tabel}. With the recent advances in pre-trained language models, table encoders adapt pre-trained language models with additional table-specific modules encoding structure~\cite{herzig2020tapas,yin2020tabert,wang2021tuta} and numeracy~\cite{wang2021tuta,herzig2020tapas}. 
These methods are intentionally built for tables, but their necessity in each task remains unknown. Our work exploits a generic model to show that content-emphasized tasks like retrieval do not require such specific designs.

\paragraph{Table Retrieval} 
Earlier works focus on web table search in response to keyword queries~\citep{cafarella2008webtables,cafarella2009data,balakrishnan2015applying,pimplikar2012answering} or a seed table~\citep{sarmad2012finding}. Many of them use the 60 keywords and relevant web tables collected by~\citet{zhang2018ad}. Tables are modeled by aggregating multiple fields~\cite{zhang2019table2vec}, contexts~\cite{trabelsi2019improved}, and synthesized schema labels~\cite{chen2020leveraging}. 
More recently, ~\citet{chen2020table,wang2021retrieving} use structure-augmented BERT for retrieval.
These works largely treat the retrieval task on its own account and target similarity under the traditional Information Retrieval (IR). 

\section{Conclusion}
Given the importance of finding relevant tables when answering questions in the NQ-table dataset, we study the task of table retrieval and find that table retrieval emphasizes content rather than table structure. Our experiments with the text-generic Dense Passage Retriever (DPR) and the state-of-the-art table-specific Dense Table Retriever (DTR) demonstrate that DPR can already encode simple structures based on linearized tables and table-specific designs such as auxiliary embeddings, hard attention mask, and soft attention bias are not necessary. Our findings suggest that future development on table retrieval can potentially be built upon successful text retrievers and table-specific model designs should be carefully examined to avoid unnecessary complexity.

\section*{Acknowledgements}
We would like to thank Frank F. Xu and Kaixin Ma for the helpful discussions and anonymous reviewers for their valuable suggestions on this paper.

\bibliography{anthology,custom}
\bibliographystyle{acl_natbib}


\end{document}